\documentclass{article}

\usepackage{microtype}
\usepackage{graphicx}
\usepackage{subfigure}
\usepackage{booktabs} % for professional tables

\usepackage{hyperref}

\usepackage{arxiv2025}

\mlsystitlerunning{FP8-Flow-MoE}

\usepackage{amssymb}
\usepackage{amsmath} % Required for some math elements 

\usepackage{algorithmic}
\usepackage{array}
\usepackage{lipsum}
\usepackage{hyperref}
\usepackage{soul}
\usepackage{enumitem}

\usepackage{booktabs}
\usepackage{multirow}
\usepackage{makecell}
\usepackage{array}
\usepackage{graphicx} 
\usepackage{subcaption}
\usepackage{dblfloatfix}
\usepackage{cuted}
\usepackage{capt-of} % 提供 \captionof

\begin{document}

\twocolumn[
\mlsystitle{FP8-Flow-MoE: A Casting-Free FP8 Recipe without Double Quantization Error}

%\mlsyssetsymbol{equal}{*}

\begin{mlsysauthorlist}
\mlsysauthor{Fengjuan Wang}{zj}
\mlsysauthor{Zhiyi Su}{zj}
\mlsysauthor{Xingzhu Hu}{zj}
\mlsysauthor{Cheng Wang}{zj}
\mlsysauthor{Mou Sun}{zj}
\end{mlsysauthorlist}

\mlsysaffiliation{zj}{Zhejiang Lab}

\mlsyscorrespondingauthor{Mou Sun}{123sssmmm@gmail.com}

%\mlsyskeywords{Machine Learning, MLSys}

\vskip 0.3in

\printAffiliationsAndNotice{}  % leave blank if no need to mention equal contribution
%\printAffiliationsAndNotice{\mlsysEqualContribution} % otherwise use the standard text.

\begin{abstract}

Training large Mixture-of-Experts (MoE) models remains computationally prohibitive due to their extreme compute and memory demands. Although low-precision training promises to accelerate computation and reduce memory footprint, existing implementations still rely on BF16-dominated dataflows with frequent quantize–dequantize (Q/DQ) conversions. These redundant casts erode much of FP8’s theoretical efficiency. However, naively removing these casts by keeping dataflows entirely in FP8 introduces double quantization error: tensors quantized along different dimensions accumulate inconsistent scaling factors, degrading numerical stability.

We propose FP8-Flow-MoE, an FP8 training recipe featuring a quantization-consistent FP8-centric dataflow with a scaling-aware transpose and fused FP8 operators that streamline computation and eliminate explicit cast operations from 12 to 2. Evaluations on a 671B-parameter MoE model demonstrate up to 21\% higher throughput and 16.5~GB lower memory usage per GPU compared to BF16 and naïve FP8 baselines, while maintaining stable convergence. We provide a plug-and-play FP8 recipe compatible with TransformerEngine and Megatron-LM, which will be open-sourced soon. %which will be open-sourced after the camera-ready release of this paper.

\end{abstract}

]

%\printAffiliationsAndNotice{\mlsysEqualContribution} % otherwise use the standard text.

%----------------------------------------------------------------------------------------
%	Main Part
%----------------------------------------------------------------------------------------
\section{Introduction}
\label{sec:introduction}

The development of large language models (LLMs) has witnessed remarkable progress in recent years, with model scales growing to hundreds of billions and even trillions of parameters~\cite{brown2020language,2022arXiv220501068Z,2023arXiv230709288T}. These massive models have demonstrated unprecedented capabilities across natural language understanding, reasoning, and knowledge-intensive tasks~\cite{openai2023gpt4,2023arXiv230312712B}. However, training such large-scale models poses significant computational challenges, and the enormous cost has become a de facto barrier for many research institutions and even large industrial labs.

To mitigate this, the Mixture-of-Experts (MoE) architecture~\cite{lepikhin2021gshard,fedus2022switch,rajbhandari2022deepspeedmoe} has been introduced to increase model capacity without proportionally increasing computation, activating only a subset of experts per token. While MoE significantly reduces the overall training cost compared to dense transformers, its large communication footprint and memory demand still make training at scale highly expensive—motivating the exploration of low-precision formats such as FP8 to further improve efficiency.

Low-precision computation has emerged as a promising direction to address this challenge. Compared to the mainstream BF16-based Automatic Mixed Precision (AMP) scheme, adopting the FP8 format can theoretically double computation throughput and halve communication latency~\cite{2022arXiv220905433M}. Since the introduction of NVIDIA’s Hopper architecture with native FP8 Tensor Core support~\cite{nvidia2022hopper}, FP8 training has drawn increasing attention. DeepSeek first demonstrated the feasibility of FP8 precision at massive scale in its V3 Mixture-of-Experts (MoE) model~\cite{ds2024v3}, and subsequently open-sourced \textit{DeepGEMM}~\cite{deepgemm2025} and \textit{DeepEP}~\cite{deepep2025}—two critical kernels for FP8 MoE efficiency—enabling high-throughput grouped GEMM computation and low-latency cross-node all-to-all communication.

Despite these advances, a complete training recipe designed around an FP8 dataflow has been missing. Naively replacing BF16 kernels with their FP8 counterparts introduces \textbf{excessive casting overheads}—each GEMM boundary typically involves multiple quantize–dequantize (Q/DQ) conversions~\cite{ds2024v3,nvidia2023transformerengine,ds2025insight}. As a result, the realized throughput of current FP8 systems often falls short of their theoretical potential, sometimes even lagging behind optimized BF16 baselines. Moreover, without a properly designed recipe and careful memory management, FP8 kernels may incur extra activation copies, undermining the expected memory savings.

However, simply removing these casts is not a viable solution. A straightforward strategy is to keep operator inputs and outputs directly in FP8—bypassing BF16 storage and boundary casting—to avoid Q/DQ overhead. While this approach reduces conversions, it introduces \textbf{double quantization error}: tensors quantized along different dimensions (e.g., row-wise before GEMM and column-wise afterward) accumulate inconsistent scaling factors, leading to directional precision loss and potential instability~\cite{NEURIPS2022_c3ba4962}.

To address these challenges, we propose \textbf{FP8-Flow-MoE}, a systematic and practical FP8 training recipe centered on an end-to-end FP8 dataflow. Our key contributions are summarized as follows:

\begin{enumerate}
    \item \textbf{Avoidance of double quantization error in FP8 dataflows.}
    We eliminate quantization inconsistency by introducing a \textbf{scaling-aware transpose} operator that directly converts FP8 tensors between row- and column-wise layouts while preserving scaling consistency. This design removes the need for dequantize–requantize cycles across operators, effectively avoiding double quantization error in FP8 dataflows.
    
    \item \textbf{Casting-free FP8 recipe with a native FP8 kernel suite.}
    We present the FP8-Flow-MoE recipe, where the entire MoE stage operates in FP8 with no explicit casting except at the entry point. Compared to a drop-in kernel replacement approach, the number of cast operations is reduced from 12 to 2, marking a paradigm shift from BF16-centric to FP8-centric computation. We further provide a suite of native FP8 operators that make the recipe feasible and efficient in practice. Several of these kernels have already been merged into upstream repositories (e.g., NVIDIA/TransformerEngine~\cite{nvidia2023transformerengine}), while others will be released as part of our open-source implementation.
    
    \item \textbf{Comprehensive analysis of FP8 training at system scale.}
    We conduct an in-depth study comparing FP8 and BF16 across MoE components, analyzing both kernel-level and quantization-related effects. FP8-Flow-MoE achieves up to 21\% throughput improvement and 7.4~GB memory reduction per GPU compared to the BF16 baseline, while naive FP8 kernel replacement yields only an 3\% gain with negligible or even negative memory savings.
\end{enumerate}

The remainder of this paper is organized as follows.  
Section~\ref{sec:related-works} reviews related work on low-precision training techniques for LLMs and the architectural design of Mixture-of-Experts models.  
Section~\ref{sec:fp8-flow} details how the scaling-aware transpose operator eliminates double-quantization error and enables a casting-free FP8 dataflow, supported by a suite of high-performance FP8 kernels integrated into FP8-Flow-MoE.  
Section~\ref{sec:experiments} compares FP8-Flow-MoE, BF16, and naïve FP8 on a 671B DeepSeek-V3 model for compute, memory, and communication efficiency, and validates convergence on a 16B DeepSeek-V2-lite model trained with 200B tokens.  
Finally, Section~\ref{sec:conclusions} concludes the paper and discusses potential directions for extending FP8-Flow-MoE to other low-precision formats and representations.

\section{Background and Related Work}
\label{sec:related-works}

\subsection{Low-Precision Training and FP8 Quantization}

Low-precision computation has become a key enabler for efficient large-scale model training. Early FP16 and BF16 mixed-precision schemes significantly reduced training cost while maintaining convergence~\cite{micikevicius2018mixed, wang2018training}. 
Recently, the FP8 format has gained attention for its ability to double arithmetic throughput on modern accelerators~\cite{nvidia2022hopper, 2022arXiv220905433M, ds2024v3}. 
FP8 generally follows IEEE 754 conventions but defines multiple variants, most notably E4M3 and E5M2, which balance precision and dynamic range differently~\cite{wang2018training}. 
E4M3 offers finer precision, while E5M2 provides a wider numeric range.  
Another variant, UE8M0, encodes powers of two and is typically used for scaling factors rather than activations.

Converting higher-precision tensors (e.g., FP32 or BF16) into FP8 requires quantization. 
The scaling strategy defines the quantization recipe and directly impacts stability and accuracy. 
Early works relied on per-tensor scaling~\cite{micikevicius2018mixed}, which often underutilizes FP8’s range when tensor values span several magnitudes. 
Later studies proposed finer-grained scaling, such as per-block or per-channel quantization~\cite{peng2023fp8}, which preserve accuracy for small-magnitude values while maintaining efficiency. 
These techniques have become standard in recent FP8 training systems.

\subsection{FP8 Training Systems and MoE Architectures}

The adoption of FP8 was accelerated by both hardware and software advances. 
NVIDIA’s Hopper architecture introduced native FP8 Tensor Core support~\cite{nvidia2022hopper}, while TransformerEngine provided open-source operator-level support for FP8 quantization~\cite{nvidia2023transformerengine}. 
Early large-scale studies, including simulated FP8 GPT-3 training~\cite{2022arXiv220905433M} and Inflection-2~\cite{Inflection2_2023}, demonstrated that FP8 can maintain convergence comparable to BF16. 
DeepSeek-V3~\cite{ds2024v3} further validated FP8 training at trillion-parameter scale and released two key libraries—\textit{DeepGEMM}~\cite{deepgemm2025} for grouped-GEMM and \textit{DeepEP}~\cite{deepep2025} for cross-node communication—establishing a strong FP8 baseline for Mixture-of-Experts (MoE) models.  

However, these implementations still follow BF16 dominated dataflows, requiring frequent quantize-dequantize (Q/DQ) conversions across GEMM boundaries~\cite{ds2025insight}
Such redundant casting limits realized FP8 performance and increases memory traffic.  
Conversely, removing these casts introduces double quantization error, as tensors quantized along different dimensions (row-wise vs. column-wise) accumulate inconsistent scaling factors~\cite{NEURIPS2022_c3ba4962}. 
This trade-off between computational efficiency and numerical stability remains a major challenge in current FP8 systems.

\subsection{System-Level Challenges in Mega-Scale MoE Training}

The rapid scaling of large language models (LLMs)~\cite{brown2020language,2023arXiv230709288T} has pushed training infrastructure to its limits. 
Parallelization strategies such as tensor, pipeline, and data parallelism~\cite{rajbhandari2022deepspeedmoe}, together with the Mixture-of-Experts (MoE) architecture~\cite{lepikhin2021gshard,fedus2022switch}, have become essential for trillion-parameter training. 
MoE reduces active computation by routing each token to a small subset of experts, achieving high model capacity without proportional cost. 
However, the sparse and irregular computation pattern introduces new bottlenecks: all-to-all expert communication, activation padding, and dynamic routing overheads~\cite{rajbhandari2022deepspeedmoe, ds2024v3}.  
Integrating FP8 precision into such workflows is particularly challenging, as each communication or expert boundary typically requires additional quantization and data reformatting.  
These factors fragment computation into many small kernels and diminish arithmetic intensity, preventing FP8’s theoretical benefits from being fully realized in end-to-end MoE training.  

These challenges motivate our FP8-Flow-MoE design—a quantization-consistent, casting-free FP8 dataflow that eliminates redundant Q/DQ operations and maintains numerical stability throughout the MoE pipeline.
\newcommand{\norm}[1]{\lVert #1 \rVert}

\section{Method}
\label{sec:fp8-flow}

There have been extensive studies on applying FP8 quantization in transformer-based model training, particularly, dense models such as BERT, GPT, etc., achieving significant performance gain with negligible accuracy loss. This contribution focus on another archtype of models, Mixture of Experts (MOE) models, which has emerged as the most performant LLM because its high capability and reduced resource required for inference, i.e. only small parts of all experts are activated.

\subsection{Scaling-aware transpose}
\begin{figure*}[!htbp]
    \centering
    % 第一行子图 (a)
    \vspace{3em} 
    \begin{subfigure}
        \centering
        \includegraphics[width=0.9\linewidth]{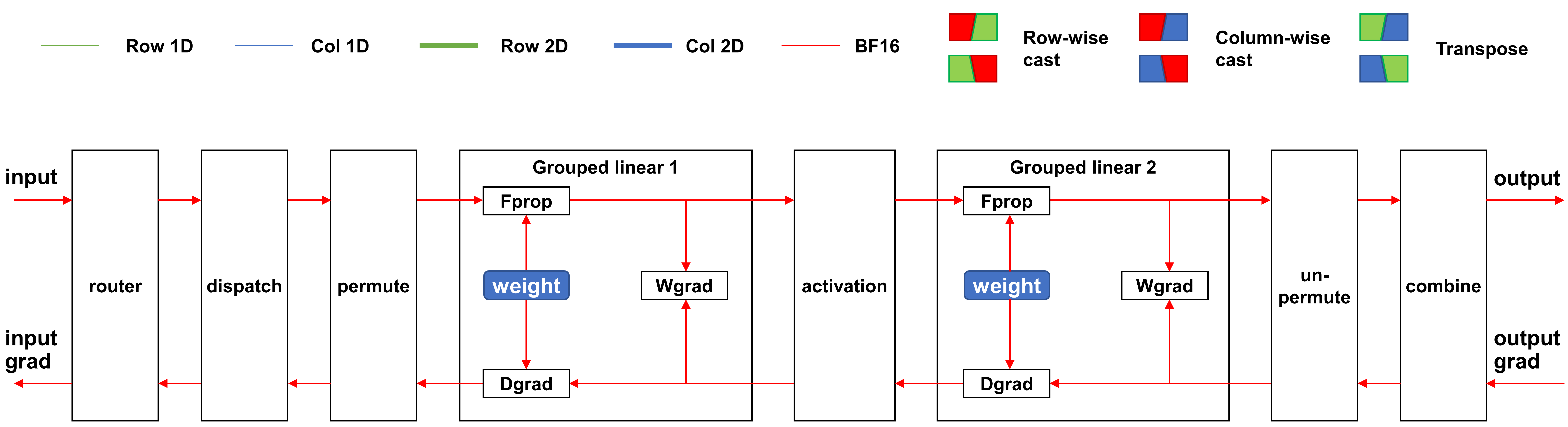}
        % \caption{The computational graph of an MoE layer with TransformerEngine blockwise recipe.}
        \caption*{(a) BF16 baseline.}
        \label{fig:experts_bf16}
    \end{subfigure}
    \vspace{2em}

    % 第二行子图 (b)
    \begin{subfigure}
        \centering
        \includegraphics[width=0.9\linewidth]{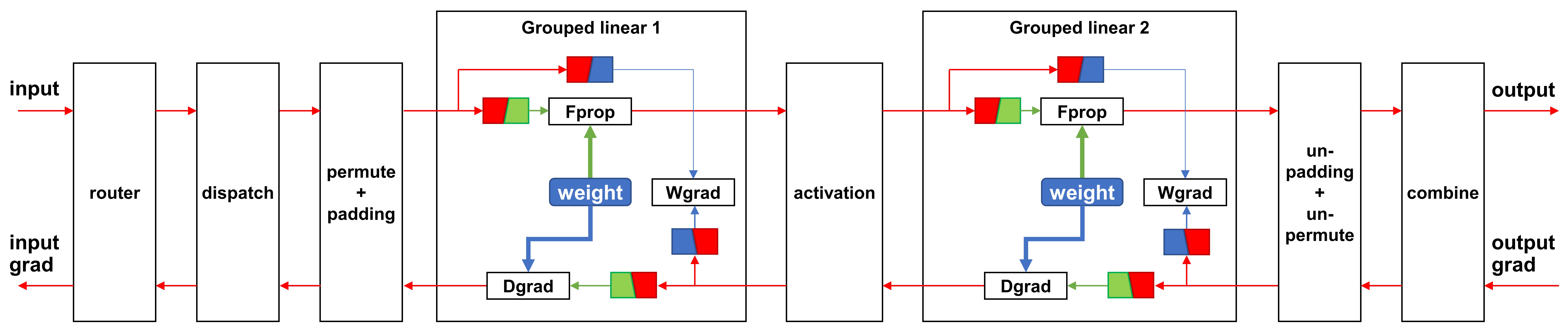}
        % \caption{FP8-Flow dataflow structure showing scaling-aware transpose.}
        \caption*{(b) TransformerEngine blockwise recipe.}
        \label{fig:blockwise_base}
    \end{subfigure}
    \vspace{2em}

    % 第三行子图 (c)
    \begin{subfigure}
        \centering
        \includegraphics[width=0.9\linewidth]{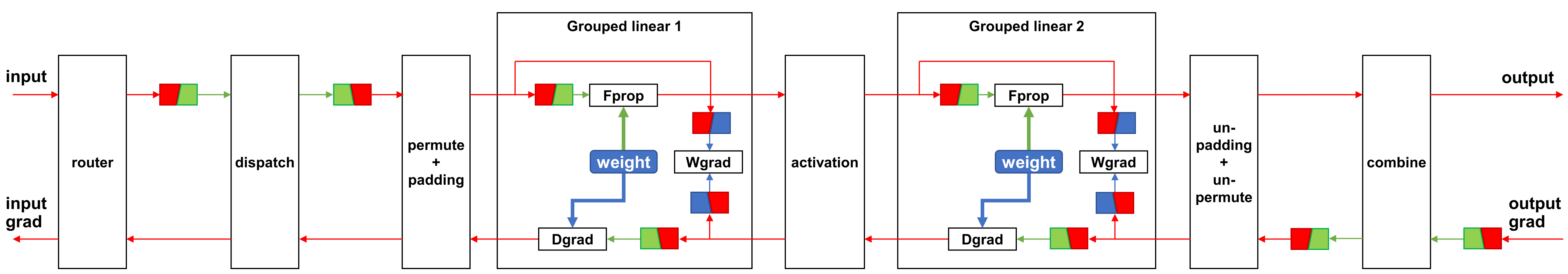}
        % \caption{Fused FP8 operator kernels in FP8-Flow.}
        \caption*{(c) according to DeepSeek V3 technical report.}
        \label{fig:dsv3_moe}
    \end{subfigure}
    \vspace{2em}

    % 第四行子图 (d)
    \begin{subfigure}
        \centering
        \includegraphics[width=0.9\linewidth]{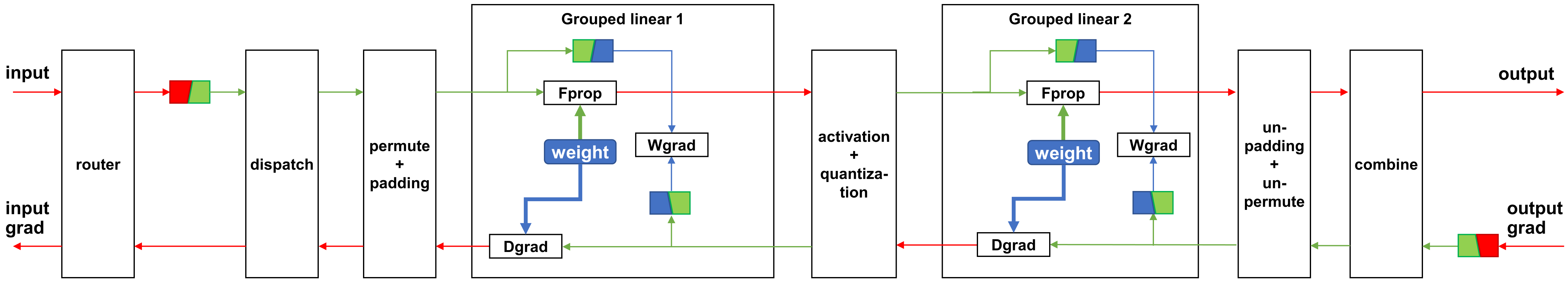}
        % \caption{Comparison of data movement and quantization overhead between FP8-Flow and baseline.}
        \caption*{(d) FP8-Flow-MoE.}
        \label{fig:fp8-flow}
    \end{subfigure}
    \vspace{2em} 
    \caption{The computational graph of an MoE module}
    \label{fig:fp8flow_overview}
\end{figure*}

One of the key challenges in fully adopting FP8 precision in the MoE module is the two different quantization layouts required by the \textit{grouped linear} operations. Specifically, the activations (and their gradients) are consumed in a row-wise quantized format (refered to as \textit{Fprop} and \textit{Dgrad} in Figure~\ref{fig:fp8flow_overview}), while the weight-gradient computation requires column-wise quantized inputs (\textit{Wgrad}). However, the preceding all-to-all communication (\textit{dispatch}) typically transmits only one format, usually the row-wise quantized data, to leverage the reduced-communication advance brought by using a low-precision format.

As a consequence, when the grouped linear layer subsequently needs column-wise quantized data, a naive implementation must first dequantize $\rightarrow$ transpose $\rightarrow$ requantize, introducing two rounds of (de)quantization. This process leads to what we call the double quantization error, whose root cause is that 1D per-token quantization uses scaling factors computed over contiguous segments (e.g., 128 elements per scale), and the scales differ across the two layouts. Since FP8 quantization maps scaled values onto a discrete representable grid determined by the exponent and mantissa bits, performing two independent quantizations with different scaling factors effectively remaps values twice to non-overlapping discrete sets. Formally, the double quantization error is defined by 

\begin{equation}
E = Q_{col}(D(Q_{row}(X)) - Q_{col}(X)  
\end{equation}

where $Q_{row}(\cdot)$, $Q_{col}(\cdot)$ are the row-wise and column-wise quantization operator and $D(\cdot)$ is the de-quantization operator. Both row-wise and column-wise quantization are performed in a per-tile fashion, with 128 continuous elements in each tile. in other words, the scaling factor is calculated as 

% Assuming the an input tensor in BF16 format, $X \in \mathbb{R}^{m \times n}$, in FP8 flow it was first quantized in a rowwise per-tile fashion, $X \in \mathbb{R}^{m \times (K \times 128)}$, where $K=n/128$.

\begin{equation}
    s = \frac{\max_{0 \leq i < 128} \norm{x_i}}{448}
\end{equation}

where 448 is the maximum representable number of FP8\_E4M3 format. After scaling, a quantization operator essentially maps $x_i$ into the nearest discrete value 

\begin{equation}
    Q_{row}(x_i) = \mathrm{round} ( \frac{x_i}{s} )
\end{equation}

A de-quantization operator scales the FP8 data back to its original range by multiplying the scaling factor, but the rounding error remains,

\begin{equation}
    D(Q_{row}(x_i)) = \mathrm{round} ( \frac{x_i}{s} ) \cdot s
\end{equation}

Following this representation, it is trivial that performing a second row-wise quantization do not introduced any extra error since the same 1x128 elements are again tiled for quantization, which gives same maximum value and eventually leaving $s$ unchanged. An number is strictly mapped to the identical discrete value, or formally,

\begin{align}
    Q_{row}(D(Q_{row}(x_i))) & = \mathrm{round} ( \frac{\mathrm{round} ( \frac{x_i}{s} ) \cdot s}{s} ) \cdot s \\
    & = \mathrm{round} ( \mathrm{round} ( \frac{x_i}{s} ) ) \cdot s \\
    & = \mathrm{round} ( \frac{x_i}{s} ) \cdot s \\
    & = Q_{row}(x_i)
\end{align}

where we assumed a deterministic rounding algorithm, e.g. round to nearest (RtN), is used and rounding the same number twice does not change its value.

However, in case that the second quantization is performed column-wise, its scaling factor will be calculated from the 128-element tile after transpose, marked as $s'$, which is generally not the same as $s$. We have 

\begin{equation}
\label{eqn:double_quant_error}
    Q_{col}(D(Q_{row}(x_i))) = \mathrm{round} ( \frac{\mathrm{round} ( {x_i}/{s} ) \cdot s}{s'} ) \cdot s' 
\end{equation}

where the two rounding operators cannot be combined because they are applying to two different values. However, This error amplification can be mitigated if the scaling factor is constrained to powers of two, in which case, rescaling between two quantization domains (row-wise and column-wise) involves only adjusting exponent bits, provided that no overflow or underflow occurs. Formally, let us assume 

$s =2^T, s'=2^{T'}$, where $T \in \mathbb{N}, T' \in \mathbb{N}$.

At row-wise quantization, the value of an element is discretized to
\begin{equation}
    Q_{row}(x) = (-1)^{SN}\cdot 2^{E-7} \cdot (1+M/8) \cdot 2^T
\end{equation}
where $SN$, $E$ and $M$ are the sign bit, exponent bits and matissa bits of the its FP8 coding respectively, 7 and 8 are constants derived from FP8\_E4M3 definition. While applying column-wise quantization, the same value is discretized to a different representation,
\begin{equation}
    Q_{col}(x) = (-1)^{SN'}\cdot 2^{E'-7} \cdot (1+M'/8) \cdot 2^{T'}
\end{equation}

Since both $T$ and $T'$ are natural numbers, there exist another natural number $D$, where $D=T'-T$. Notice that if we set 
\begin{align*}
    SN'&=SN \\
    E' &= E - D \\
    M' &= M  
\end{align*}, we have 
\begin{align}    
    Q_{col}(x) &= (-1)^{SN}\cdot 2^{E-D-7} \cdot (1+M/8) \cdot 2^{D+T} \\
    &= (-1)^{SN}\cdot 2^{E-D-7+D+T} \cdot (1+M/8) \\
    &= (-1)^{SN}\cdot 2^{E-7+T} \cdot (1+M/8) \\
    &= (-1)^{SN}\cdot 2^{E-7} \cdot (1+M/8) \cdot 2^T \\
    &= Q_{row}(x)\\    
\end{align}

Building on this insight, we further align all scaling factors within a transposed block to the largest one (to avoid overflow) and directly convert between layouts by exponent manipulation alone. In other words, we can transform row-wise quantized FP8 tensors into column-wise quantized ones without any dequantization or re-quantization, by simply modifying the exponent bits of the FP8 encodings. This principle underlies our Direct Transpose operator, as outlined in Algorithm~\ref{alg:fp8_transpose}.

\begin{algorithm}[htb]
    \caption{Scaling-aware FP8 Transpose}
    \label{alg:fp8_transpose}
\begin{algorithmic}
    \STATE {\bfseries Input:} row-wise data $X_{i,j}^{row}$, scaling factor $S_{i,j}^{row}$
    \STATE Initialize column-wise data $X_{i,j}^{col} = X_{j,i}^{row}$
    \FOR{each $128\times 128$ block}
        \STATE $S_{max}=\mathrm{\max}_{0\leq i <128} S_{i,}^{row}$
        \STATE $S_{i,}^{col}=S_{max}$
        \FOR{each element $X_{i,j}^{col}$ in block}
            \STATE calculate $SN$, $E$ and $M$ from $X_{i,j}^{col}$ 
            \STATE $k = \log_2(S_{\max})/S_{i,}^{col}$
            \STATE $E_{\mathrm{new}} = E - k$
            \STATE reassemble $X_{i,j}^{col}$ coding using $SN$, $E_{\mathrm{new}}$ and $M$
        \ENDFOR
    \ENDFOR
\end{algorithmic}
\end{algorithm}
% \REQUIRE FP8 tensor $X_{\text{rowwise}}$, scaling factors $S_{\text{rowwise}}$ (power-of-two)
%     \ENSURE FP8 tensor $X_{\text{colwise}}$, scaling factors $S_{\text{colwise}}$
%     \vspace{0.5em}
%     \FOR{each column block $C$ in $X_{\text{rowwise}}^{\top}$}
%         \STATE $S_{\max} \leftarrow \max(S_{\text{rowwise}}$ within block$)$
%         \FOR{each element $x_{ij}$ in block}
%             \STATE $k \leftarrow \log_2(S_{\max} / S_{\text{rowwise}}[i])$
%             \STATE $X_{\text{colwise}}[j,i].\text{exponent} \leftarrow X_{\text{rowwise}}[i,j].\text{exponent} - k$
%         \ENDFOR
%         \STATE $S_{\text{colwise}}[j] \leftarrow S_{\max}$
%     \ENDFOR

To validate the efficiency benefits of this operator, we conducted numerical experiments at various input dimension. For each, we compared two strategies to acquire the column-wise quantized data:

\begin{enumerate}[label=(\arabic*)]
    \item \textbf{Naive conversion}: dequantize $\rightarrow$ transpose $\rightarrow$ quantize method.
    \item \textbf{Direct Transpose (ours)}: exponent bits manipulation without (de)quantization.
\end{enumerate}

\begin{figure}[!htbp]
    \centering
    \includegraphics[width=\linewidth]{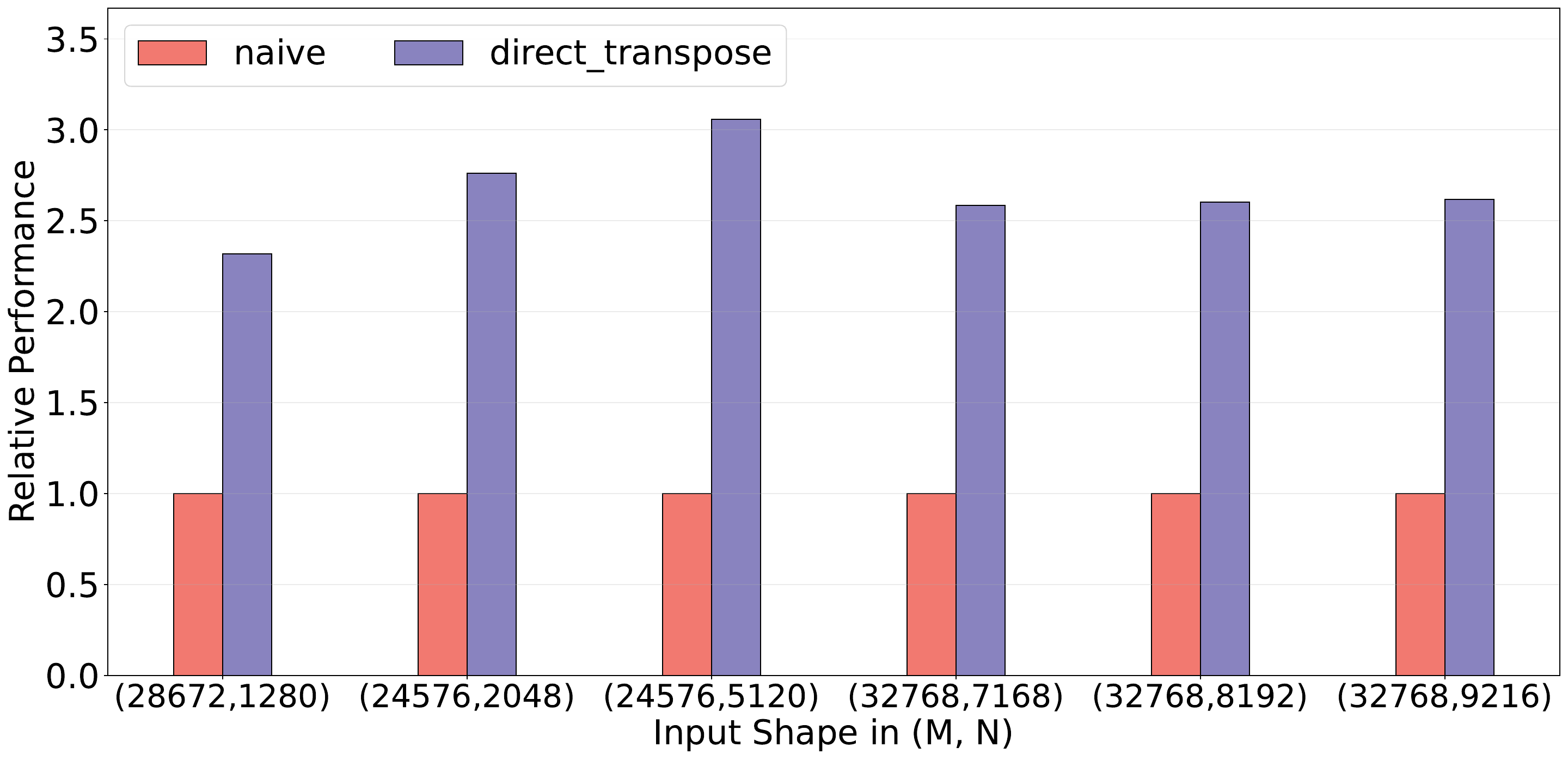}
    \caption{Comparison of latency using different quantization strategies.}
    \label{fig:latency_comp}
\end{figure}

The latency results are illustrated in Figure~\ref{fig:latency_comp}, 
% Empirically, method (1) yields substantial quantization error, while methods (2) and (3) achieve nearly identical numerical accuracy—reducing the error magnitude by over three orders compared to (1). Furthermore
it is clear that our Direct Transpose operator delivers 2 to 3 times speedup across all tensor shapes, confirming its effectiveness in computational efficiency.

% The maximum double quantization error, $\mathrm{max}(\norm{Q_{col}(X)-Q_{row}(X)})$, are computed, normalized to the direct quantization error, $\mathrm{max}(\norm{Q_{row}(X)-X})$, and summarized in table \ref{fig:latency_comp}.

\subsection{Casting-free FP8 dataflows}
The MoE architecture has emerged as an effective approach to scale model capacity without proportionally increasing the computational footprint. Within each MoE layer, a router dynamically assigns input tokens to a small subset of experts, where each expert typically consists of a two-layer feed-forward network. From a dataflow perspective, the computation of a single MoE layer can be decomposed into the following stages: routing, dispatch, permutation, expert computation, unpermutation, and combination. The expert computation itself usually includes two grouped linear projections separated by a nonlinearity (e.g., SwiGLU). In large-scale implementations, these projections are executed as grouped GEMM operations to exploit hardware efficiency across multiple experts.

Under conventional BF16-based training, the end-to-end dataflow involves multiple data format conversions across these stages. Tokens are first routed and dispatched through an all-to-all communication primitive, then permuted and packed into contiguous expert batches for efficient GEMM computation. Each expert executes the first grouped linear transformation, followed by activation, a second grouped linear transformation, and finally the unpermutation and combination steps to aggregate the outputs. This structure defines the core of the MoE layer’s computational graph.

% \begin{figure*}
%     \centering
%     \includegraphics[width=0.9\linewidth]{images/experts_bf16.png}
%     \caption{The computational graph of an MoE layer in BF16.}
%     \label{fig:experts_bf16}
% \end{figure*}

When extending this flow to FP8 computation, existing frameworks primarily adopt conservative integration strategies. For instance, NVIDIA’s Transformer Engine implements a blockwise FP8 recipe, where quantization and computation in FP8 are confined strictly within the grouped\_linear modules. This design allows FP8 acceleration in matrix multiplications (grouped GEMMs), achieving higher arithmetic throughput compared to BF16, but leaves all communication and data movement stages—particularly the dispatch all-to-all operations—in higher precision formats. As a result, while the GEMM kernels benefit from FP8 tensor cores, the overall MoE layer performance remains bounded by high-precision communication and frequent format conversions.

% \begin{figure*}
%     \centering
%     \includegraphics[width=0.9\linewidth]{images/blockwise.png}
%     \caption{The computational graph of an MoE layer with TransformerEngine blockwise recipe.}
%     \label{fig:blockwise_base}
% \end{figure*}

Another representative implementation is the DeepSeek-V3 training system, which demonstrated successful large-scale FP8 MoE training with state-of-the-art performance. Although the complete training framework was not fully open-sourced, the available descriptions and accompanying releases—specifically DeepGEMM and DeepEP—suggest that both expert computation and inter-expert communication were accelerated using customized, high-performance FP8 kernels. However, the overall dataflow still incurs substantial quantization overhead. Our analysis shows that such an approach introduces up to twelve quantization/dequantization operations within a single MoE forward-backward pass. Moreover, in the weight gradient computation (wgrad) stage, FP8 activations are typically quantized row-wise, dequantized, and then re-quantized in a column-wise manner. Since these are per-token quantization schemes with their scaling happens along a particular direction, the repeated quantization steps can amplify rounding and scaling errors ("double quantization error"), negatively affecting numerical stability and convergence.

% \begin{figure*}
%     \centering
%     \includegraphics[width=0.9\linewidth]{images/dsv3.png}
%     \caption{The computational graph of an MoE layer according to DeepSeek V3 technical report.}
%     \label{fig:dsv3_moe}
% \end{figure*}

To fully exploit the computational advantages of FP8 while maintaining numerical stability, we design an FP8-centric dataflow, referred to as FP8-Flow-MoE, that minimizes format conversions and preserves FP8 representation across the entire MoE expert path whenever possible. In this design, both the input activations and output gradients remain in FP8 throughout forward and backward propagation, except for two specific boundaries:
\begin{enumerate}
    \item between the output of the first \textit{grouped\_linear} and the activation function, and
    \item between the second \textit{grouped\_linear} and the combination/dispatch operation in the backward pass. 
\end{enumerate} 

These two exceptions correspond to fundamental numerical challenges for FP8 arithmetic. The first involves reduction operations (e.g., summations) that are highly susceptible to overflow under limited exponent range; the second involves nonlinear transformations (e.g., GELU or ReLU) that can amplify small quantization errors. Since these two computations are adjacent in the graph, we retain BF16 precision locally to ensure stability, while keeping all other tensors strictly in FP8 format. This design yields a near-continuous FP8 dataflow and substantially reduces quantization boundaries compared to prior FP8-in-MoE implementations.

% \begin{figure*}
%     \centering
%     \includegraphics[width=0.9\linewidth]{images/FP8-flow.png}
%     \caption{The computational graph of FP8-Flow-MoE}
%     \label{fig:fp8-flow}
% \end{figure*}

% \clearpage

\subsection{High-Performance FP8 Kernels}
To make the proposed FP8-centric computation flow practical and efficient, we further develop a comprehensive set of supporting operators specifically tailored to the data dependencies and numerical patterns within the MoE layer. These operators address key challenges that arise when applying FP8 to large-scale MoE training, including cumulative rounding and scaling errors, redundant data conversions, and schedule inefficiencies caused by fractured kernels. Together, they enable a seamless and high-throughput execution of the FP8-Flow-MoE, ensuring that the benefits of FP8 tensor cores extend beyond isolated kernels to the full end-to-end training pipeline.

To evaluate the effectiveness of these FP8 operators, we conduct a series of unit-level benchmarks that isolate each kernel’s performance and numerical behavior. The test inputs are generated through simulation, and their tensor shapes are chosen to reflect the real-world configurations of representative MoE models—DeepSeek V2-Lite, DeepSeek V2, and DeepSeek V3—under different expert parallelism (EP) settings. Specifically, These microbenchmarks allow us to directly measure raw kernel throughput, demonstrating how each operator contributes to the overall efficiency of the FP8-Flow-MoE system.

\subsubsection{Fused Permute and Padding}
In FP8 data pipeline, a \textit{padding} step is introduced to satisfy the shape constraint of FP8 GEMM kernels—each expert’s input must contain a multiple of 16 entries to fully utilize tensor cores. Before this, a \textit{permute} operation reorganizes dispatched tokens so that samples belonging to the same expert are contiguous in memory.
Executing these two light-weight data movement kernels separately incurs redundant HBM accesses and kernel launches. Since both operations are element-wise and independent, they can be naturally fused into a single kernel.

Based on these observations, we design a Fused Permute+Padding operator that directly performs expert-wise reordering and shape alignment in one pass. The fusion is implemented through a thread-block mapping scheme that dynamically computes the target offset in the padded layout while streaming reordered input elements from global memory. Similarly, in the backward propagation phase, we apply the same fusion strategy to combine unpermute and unpadding operations into a single kernel. This symmetric fusion ensures consistent data handling between forward and backward flows.

Figure~\ref{fig:fused_permute_padding} reports the performance comparison between the fused and unfused implementations under both BF16 and FP8 precision settings. Across different tensor sizes corresponding to typical MoE workloads, the fused version demonstrates a consistent acceleration. For the forward permute+padding case, we observe up to 1.7× speedup, while for the backward unpermute+unpadding kernel, the improvement reaches as high as 6.6× on large-scale configurations.

\begin{figure}[!htbp]
    \centering
    \includegraphics[width=0.95\linewidth]{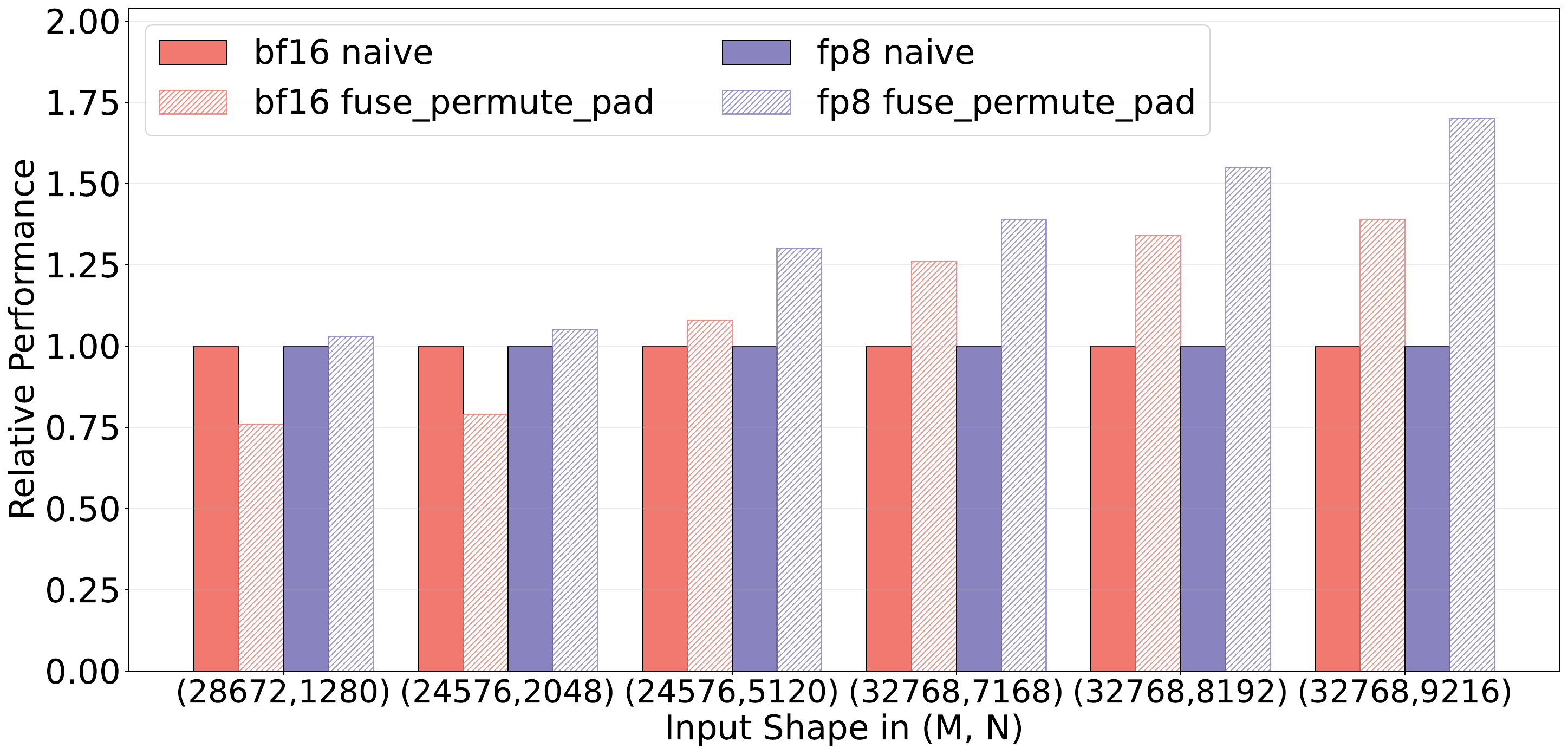}
    \caption{Performance comparison between fused and separate implementation of permute and padding kernels.}
    \label{fig:fused_permute_padding}
\end{figure}

\begin{figure}[!htbp]
    \centering
    \includegraphics[width=0.95\linewidth]{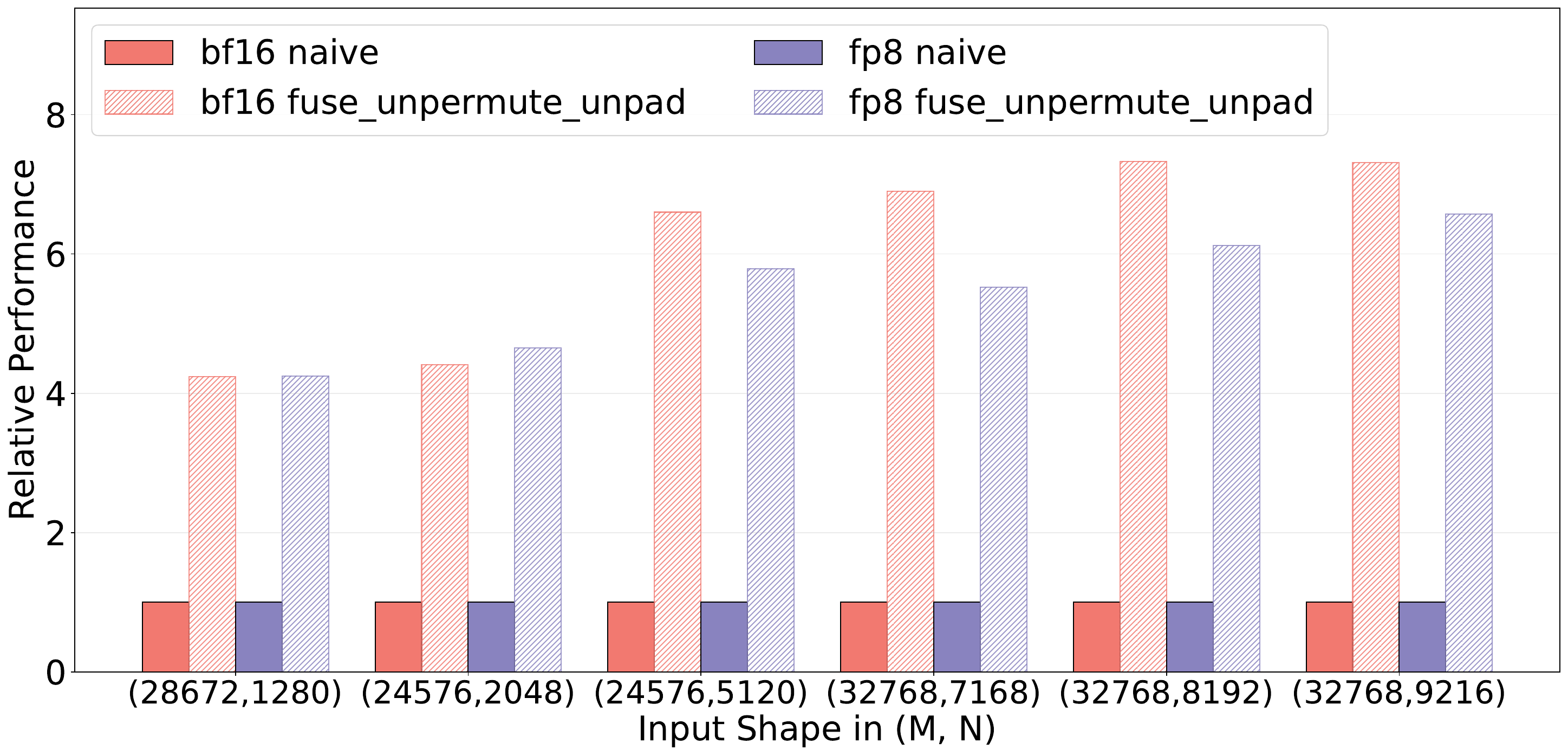}
    \caption{Performance comparison between fused and separate implementation of un-permute and un-padding kernels.}
    \label{fig:fused_permute_padding}
\end{figure}

\subsubsection{Fused SwiGLU and Quantization}
The motivation for designing an FP8-centric training flow originates from a key performance observation:
naively inserting FP8 kernels (e.g., high-performance GEMM and communication) into a BF16 computation graph introduces frequent quantization and dequantization operations, which can substantially offset the benefits of FP8 acceleration.

To illustrate this issue, we analyze FP8 communication using the DeepEP library released by DeepSeek. In its default usage, quantization and dequantization are performed immediately before and after the communication. 
Table~\ref{comm-perf-detailed-table} reports the measured latency under three representative data scales (corresponding to DeepSeek V2-Lite, V2, and V3) and three expert-parallel (EP) degrees (8, 16, 32). The results reveal two key findings: for smaller-scale workloads, quantization and dequantization each consume a similar amount of time as the communication kernel, nearly eliminating any performance gain; While the communication cost grows roughly linearly with workloads, the (de)quantization cost remains constant, however, such data conversion leads to a reduction of end-to-end FP8 speedup from 1.6× to 1.4×.

It is worth noting that although FP8 communication theoretically halves the data transfer size, in practice, the need to transmit both the FP8 tensor and its corresponding scaling factors doubles the number of data buffers and synchronizations, limiting achievable speedup to around 1.6×.
Nevertheless, a single pair of quantization and dequantization operators reduces the performance gain of FP8 communication by roughly one third (from 0.6× to 0.4× improvement). Considering that a typical MoE forward or backward pass involves around three such pairs (see Figure~\ref{fig:fp8flow_overview}(c)), the benefit of FP8 kernels can be almost completely neutralized.

\begin{figure}[!htbp]
    \centering
    \includegraphics[width=0.95\linewidth]{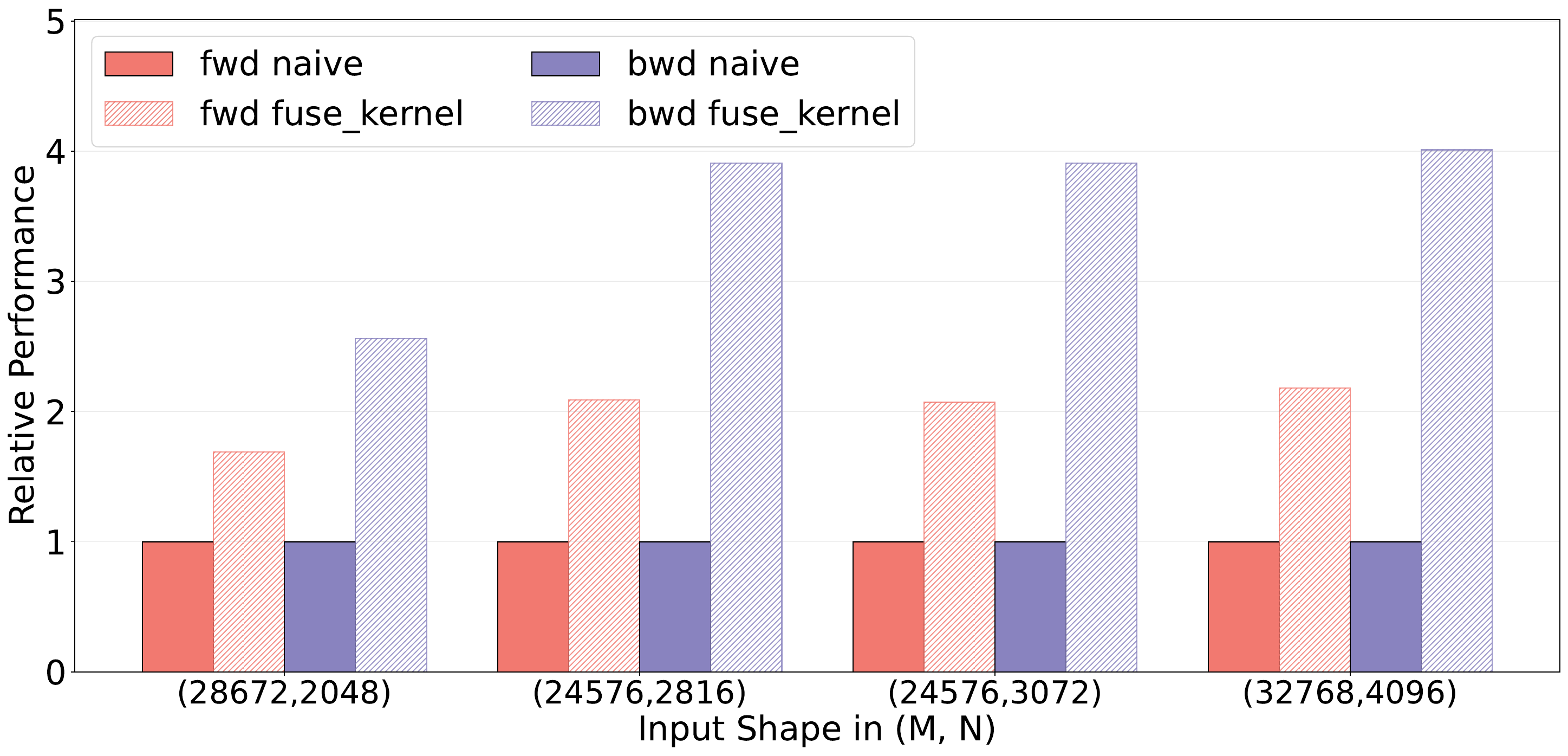}
    \caption{Performance of fused SwiGLU and quantization kernel.}
    \label{fig:fused_swiglu_quant}
\end{figure}

\begin{table}[!htbp]
\vskip 0.1in
\centering
\begin{small}
\begin{sc}
\scriptsize                      % 用更小字号节省水平空间
\setlength{\tabcolsep}{3pt} % reduce column spacing
\begin{tabular}{@{}c c c c c c c@{}}
\toprule
(M,N,EP)& BF16 & \multicolumn{3}{c}{FP8 Time (ms)} & \multicolumn{2}{c}{Speedup} \\
\cmidrule(r){3-5} \cmidrule(l){6-7}
& (ms)& Q/D & Comm & All & Comm & All \\
\midrule
(24576,2048,8)& 0.537 & 0.127/0.084 & 0.325 & 0.535 & 1.65× & 1.00× \\
(24576,5120,8)& 0.785 & 0.087/0.089 & 0.526 & 0.703 & 1.49× & 1.12× \\
(32768,7168,8)& 1.276 & 0.086/0.089 & 0.905 & 1.080 & 1.41× & 1.18× \\
(24576, 2048,16)& 1.224 & 0.091/0.083 & 1.176 & 1.350 & 1.04× & 0.91× \\
(24576,5120,16)& 2.213 & 0.082/0.082 & 1.400 & 1.564 & 1.58× & 1.42× \\
(32768,7168,16)& 2.934 & 0.084/0.092 & 1.847 & 2.023 & 1.59× & 1.45× \\
(24576,2048,32)& 3.005 & 0.094/0.083 & 2.740 & 2.918 & 1.10× & 1.03× \\
(24576,5120,32)& 5.003 & 0.082/0.081 & 2.868 & 3.031 & 1.74× & 1.65× \\
(32768,7168,32)& 7.327 & 0.082/0.082 & 4.319 & 4.483 & 1.70× & 1.63× \\
\bottomrule
\end{tabular}
\end{sc}
\end{small}
\vskip -0.05in
\caption{Communication Performance with Speedup}
\label{comm-perf-detailed-table}
\end{table}
To mitigate this issue, FP8-Flow-MoE integrates quantization directly into surrounding compute kernels whenever possible.
In particular, we fuse the SwiGLU activation and quantization steps into a single kernel. Figure~\ref{fig:fused_swiglu_quant} presents the results of our unit tests across several representative tensor shapes. The fused implementation achieves nearly identical latency to the original standalone SwiGLU operator in NVIDIA’s Megatron framework, while seamlessly producing FP8-quantized outputs for subsequent computation. This demonstrates that FP8-Flow-MoE can preserve the numerical efficiency of mixed precision while removing the quantization bottleneck that previously limited end-to-end performance.

\section{Experiments}
\label{sec:experiments}
We evaluate FP8-Flow-MoE from two perspectives: numerical convergence and training efficiency.
All experiments are conducted on a 32-node NVIDIA Hopper cluster, where each node is equipped with eight GPUs (80 GB memory per GPU).
Expert parallelism (EP) is scaled across nodes to assess training efficiency and scalability under realistic large-model workloads.

\subsection{Convergence Validation}

To verify that FP8-Flow-MoE maintains numerical stability and convergence behavior,
we train a DeepSeek-V2-Lite model (16 B parameters) from scratch under two precision settings: BF16 and FP8-Flow-MoE.
Both models are trained for 200 B tokens, using identical optimization hyperparameters, learning rate schedule, data ordering, and parallelism configurations.

\begin{figure}[!htbp]
    \centering
    \includegraphics[width=0.9\linewidth]{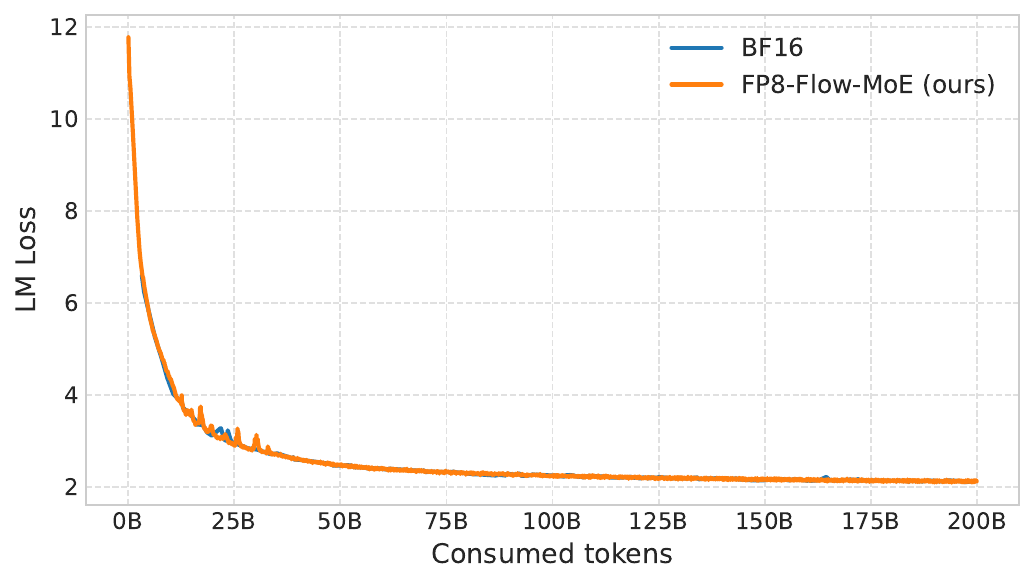}
    \caption{Training loss comparison across 200 billion tokens}
    \label{fig:convergence}
\end{figure}

As shown in Fig.~\ref{fig:convergence}, the two loss curves are nearly indistinguishable across the entire training trajectory—from the early warm-up phase through the stabilized loss regime.
The FP8-Flow-MoE model shows no sign of divergence, gradient underflow, or numerical drift, demonstrating that our fp8-centric dataflow does not compromise convergence fidelity even under extended-scale workloads.
This observation confirms that the scaling-aware transpose and fused operators in FP8-Flow-MoE preserve the numerical dynamics of BF16 training.

\subsection{Efficiency Evaluation}
We next benchmark end-to-end training efficiency across multiple precision and dataflow configurations. Specifically, we compare the following three representative configurations:
\textbf{(1)~BF16}: baseline configuration using standard FP32-BF16 mixed precision.
\textbf{(2)~Blockwise}: TE-style blockwise recipe with FP8 grouped GEMM kernels.
\textbf{(3)~FP8-Flow-MoE (ours)}: FP8-centric dataflow with fused operators and minimal quantization overhead.

% For this set of experiments, we train the DeepSeek-V2 236 B model under two activation checkpointing settings: AC=full, enabling full recomputation;
% AC=sel, disabling recomputation for MoE layers while keeping it enabled elsewhere. We further vary the expert parallelism (EP) degree across 8, 16, and 32 to observe scaling trends.

% For this set of experiments, 
All experiments are conducted on the DeepSeek-V3 (671B) model to validate the scalability and robustness of FP8-Flow-MoE under realistic large-scale training conditions. 
Models are trained using standard expert and pipeline parallelism configurations (EP/PP = 8/32, 16/16, 32/8) and evaluated under two activation checkpointing (AC) strategies: 
\textit{AC~=~full}, which applies full checkpointing to all modules; and 
\textit{AC~=~sel (+MoE expert)}, which selectively checkpoints the MoE layer while excluding experts to evaluate FP8 activation compression. 
The latter represents the most memory-efficient configuration compatible with the 1F1B-overlap schedule in Megatron, as checkpointing the entire MoE layer would otherwise cause OOM across all recipes. 
To isolate the effect of computation efficiency from communication overlap, all end-to-end experiments are performed without computation–communication overlap.
 
\textbf{Throughput Analysis.} 
Table~\ref{tab:tgs_mem_ac_full} and~\ref{tab:tgs_mem_ac_sel_moe} summarizes the measured throughput (TGS, tokens/GPU/s) and peak memory (Mem, GB) under three EP settings. Across all activation checkpointing modes, FP8-Flow-MoE consistently outperforms the BF16 baseline, achieving +6 \% (EP8), \textbf{+}8 \% (EP16), and +16 \% (EP32) throughput improvements. Compared with the Blockwise recipe, FP8-Flow-MoE further improves throughput by 3 \% (EP8), 8 \% (EP16), and up to 21 \% (EP32). The performance gap widens at higher EP levels, where the cumulative impact of reduced quantization overhead and kernel fusion becomes more significant.  

 \textbf{Memory Efficiency.} 
 Under \textit{AC = sel (+MoE expert)} at EP8, FP8-Flow-MoE reduces peak memory by approximately 8 GB vs. BF16 and 16.5 GB vs. Blockwise, directly benefiting from the FP8 checkpoint compression mechanism, which stores intermediate activations in FP8 instead of BF16. At the largest scale of EP = 32, both the BF16 and Blockwise baselines encounter out-of-memory (OOM) errors, while FP8-Flow-MoE remains stable, clearly demonstrating its superior scalability and memory efficiency. 
 
\begin{table}[!htbp]
\centering
\small
\setlength{\tabcolsep}{4pt}

\begin{tabular}{lcccccc}
\toprule
\multirow{2}{*}{\textbf{Method}} &
\multicolumn{2}{c}{\textbf{EP8}} &
\multicolumn{2}{c}{\textbf{EP16}} &
\multicolumn{2}{c}{\textbf{EP32}} \\
\cmidrule(lr){2-3}\cmidrule(lr){4-5}\cmidrule(lr){6-7}
& \textbf{TGS} & \textbf{Mem}
& \textbf{TGS} & \textbf{Mem}
& \textbf{TGS} & \textbf{Mem} \\
\midrule
\textbf{BF16}
& 1{,}109 & 39
& 939 & \textbf{36}
& 671 & \textbf{43} \\
\textbf{Blockwise}
& 1{,}146 & 37
& 938 & 41
& 644 & 51 \\
\textbf{FP8-Flow-MoE}
& \textbf{1{,}176} & 37
& \textbf{1{,}012} & 39
& \textbf{779} & 49 \\
\bottomrule
\end{tabular}
\caption{Throughput analysis under \textbf{AC=full}}
\label{tab:tgs_mem_ac_full}
\end{table}

\begin{table}[!htbp]
\centering
\small
\setlength{\tabcolsep}{4pt}

\begin{tabular}{lcccccc}
\toprule
\multirow{2}{*}{\textbf{Method}} &
\multicolumn{2}{c}{\textbf{EP8}} &
\multicolumn{2}{c}{\textbf{EP16}} &
\multicolumn{2}{c}{\textbf{EP32}} \\
\cmidrule(lr){2-3}\cmidrule(lr){4-5}\cmidrule(lr){6-7}
& \textbf{TGS} & \textbf{Mem}
& \textbf{TGS} & \textbf{Mem}
& \textbf{TGS} & \textbf{Mem} \\
\midrule
\textbf{BF16}
& 1{,}178 & 64
& 1{,}055 & 71
& OOM & OOM \\
\textbf{Blockwise}
& 1{,}178 & 73
& 1{,}031 & 77
& OOM & OOM \\
\textbf{FP8-Flow-MoE}
& \textbf{1{,}193} & \textbf{56}
& \textbf{1{,}111} & \textbf{66}
& \textbf{912} & \textbf{75} \\
\bottomrule
\end{tabular}
\caption{Throughput analysis under \textbf{AC=sel (+MoE expert)}}
\label{tab:tgs_mem_ac_sel_moe}
\end{table}

\subsection{Discussion and Insights}

The experimental results collectively highlight three key findings:

\textbf{Persistent FP8 dataflow matters}. Simply replacing GEMM or communication kernels with FP8 counterparts yields limited benefit, since quantization/dequantization costs dominate at small batch sizes or high communication frequency. The proposed scaling-aware transpose and fused operator design eliminates redundant precision transitions, directly improving kernel launch efficiency and memory bandwidth utilization.

\textbf{Scaling amplifies FP8-Flow-MoE’s gains}. As EP increases, communication and quantization overheads compound in baseline methods, whereas FP8-Flow-MoE’s low-precision persistence minimizes these costs, resulting in up to $1.2\times$ end-to-end acceleration at large expert parallalism.

\textbf{Practical training stability}. Despite adopting lower precision for a larger portion of the training pipeline than baseline implementations, FP8-Flow-MoE preserves identical convergence behavior and robustness across hundreds of billions of tokens. This stability stems from our insight of potential double quantization errors at transpose boundaries. By introducing a scaling-aware transpose, FP8-Flow-MoE effectively eliminates this redundant quantization while streamlining data movement, yielding both improved numerical fidelity and higher efficiency.

\section{Conclusions}
\label{sec:conclusions}

This study identifies two additional factors that play critical roles in achieving high end-to-end efficiency for FP8 precision in large-scale Mixture-of-Experts (MoE) training.

First, maintaining data in low precision throughout the computation flow minimizes redundant conversions and alleviates memory bandwidth pressure. A representative example is our FP8 transpose operator, which directly transforms row-wise to column-wise quantized tensors without reverting to higher precision—achieving high data throughput while avoiding double quantization error.

Second, FP8 precision inevitably fragments the computation graph by introducing small operators such as quantization and padding, leading to excessive kernel launches and memory traffic. By fusing lightweight operations and developing custom high-performance implementations, FP8-Flow-MoE effectively eliminates this structural inefficiency.

In summary, FP8-Flow-MoE establishes the first FP8-centric training paradigm for MoE models, maintaining convergence parity with BF16 while achieving up to 21\% higher throughput and 16.5~GB lower peak memory per GPU compared with BF16 workflows and naïve FP8 kernel replacements. These results demonstrate the feasibility of fully low-precision MoE training and open new opportunities for scalable, energy-efficient foundation model development.

%----------------------------------------------------------------------------------------
%	Bibliography
%----------------------------------------------------------------------------------------
\clearpage
\bibliography{main}
\bibliographystyle{arxiv2025}

%----------------------------------------------------------------------------------------
%	Appendix
%----------------------------------------------------------------------------------------
\appendix

\end{document}